%% file: main.tex
\documentclass[letterpaper, 10 pt, conference]{ieeeconf}
\usepackage{times}
\usepackage{cite}
\usepackage{multicol}
\usepackage[bookmarks=true,hidelinks]{hyperref}

\usepackage{amsmath,amssymb,amsfonts,amsthm}
\usepackage{algorithmic}
\usepackage{graphicx}
\usepackage{textcomp}
\usepackage{xcolor}
\usepackage{multirow}

\IEEEoverridecommandlockouts

\newtheorem{theorem}{Theorem}[section]

\theoremstyle{definition}
\newtheorem{definition}{Definition}[section]

\begin{document}
\title{\bf Differentiable Trajectory Generation for Car-like Robots with Interpolating Radial Basis Function Networks}

\author{Hongrui Zheng, Rahul Mangharam
\thanks{All authors are with the University of Pennsylvania, Department of Electrical and Systems Engineering, 19104, Philadelphia, PA, USA. Emails: \{\tt\small hongruiz, rahulm\}@seas.upenn.edu}
}

\maketitle

\input{abstract}
\input{intro.tex}
\input{method.tex}
\input{exp.tex}
\input{discussion.tex}
\input{ack.tex}

\bibliographystyle{IEEEtran}
\bibliography{main}

\end{document}

%% file: abstract.tex
\begin{abstract}
The design of Autonomous Vehicle software has largely followed the \textit{Sense-Plan-Act} model. Traditional modular AV stacks develop perception, planning, and control software separately with little integration when optimizing for different objectives. On the other hand, end-to-end methods usually lack the principle provided by model-based white-box planning and control strategies. We propose a computationally efficient method for approximating closed-form trajectory generation with interpolating Radial Basis Function Networks to create a middle ground between the two approaches. The approach creates smooth approximations of local Lipschitz continuous maps of feasible solutions to parametric optimization problems. We show that this differentiable approximation is efficient to compute and allows for tighter integration with perception and control algorithms when used as the planning strategy.
\end{abstract}

%% file: intro.tex
\section{Introduction}

Traditionally, the motion planning task for a car-like robot requires synthesizing trajectories online. The local planning task ultimately searches for a sequence of feasible control inputs given a desired local goal. This is naturally expressed as a parametric optimization problem. One can easily enforce constraints from vehicle dynamics or limits of operation range. However, even with efficient optimization solvers, solving potentially hundreds of optimization with different specifications online up to thirty times a second is still challenging. Massive look-up tables storing discretized solutions found by running the optimizations offline have been used to speed up the process online. However, the look-up table can only provide discrete approximations of the actual optimal control. Moreover, the trajectory generation remains single-threaded and requires high memory usage without specialized software implementation.

\begin{figure}
    \centering
    \includegraphics[width=0.9\columnwidth]{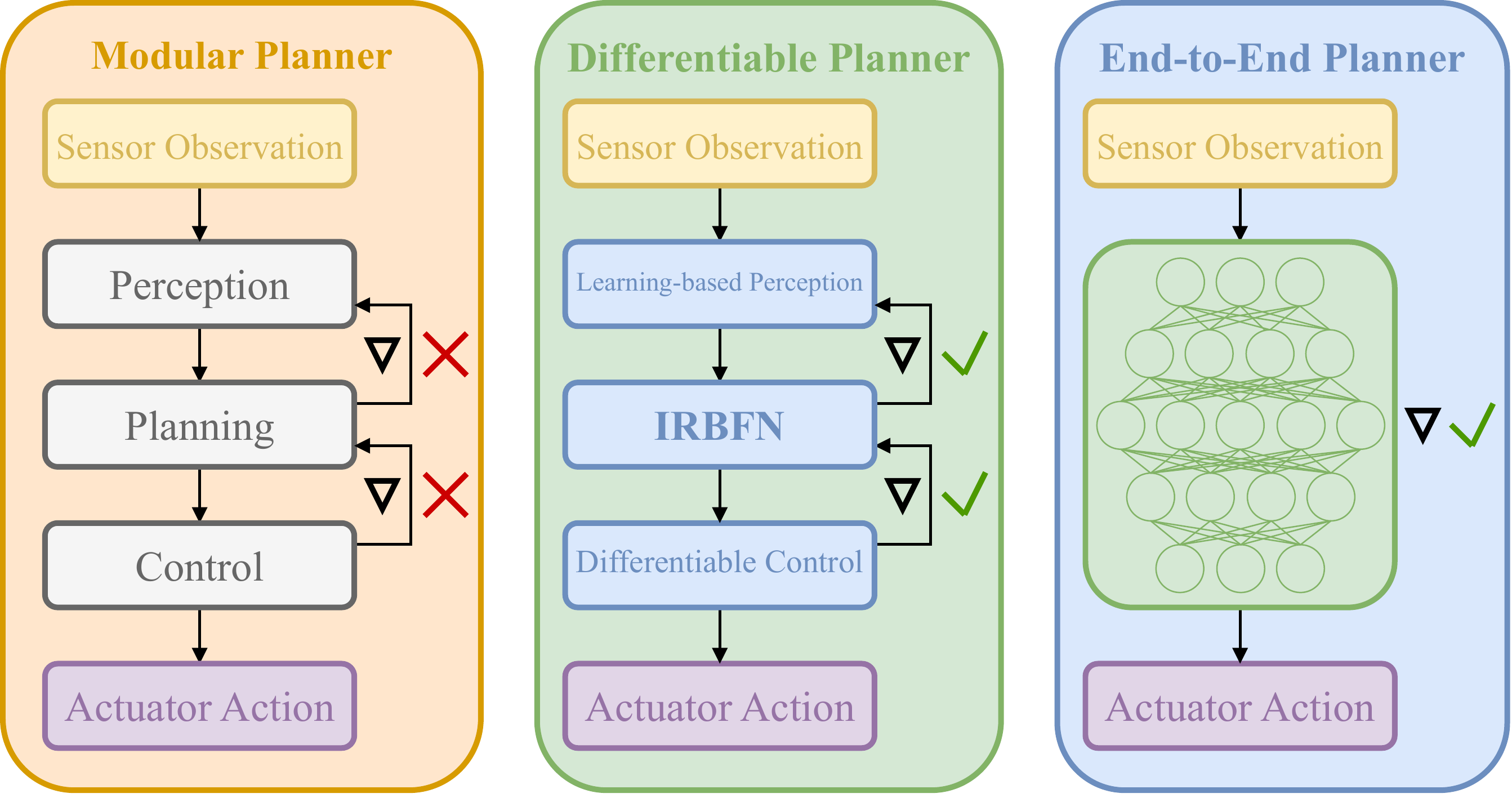}
    \caption{Comparison between standard modular planner stack, differentiable planner stack, and end-to-end planner stack. Gradient information is not available between modules of a traditional modular planner, and is available in differentiable and end-to-end planners.}
    \label{fig:stack_compare}
    \vspace{-15pt}
\end{figure}

More recently, different efforts have tried to bypass solving optimal control online by creating end-to-end planners attempting to directly produce control inputs using sensor information \cite{bojarski2016end, sallab2017deep, jain2021autonomy}. End-to-end approaches provide the benefit of exposing gradient information for upstream and downstream processes (e.g., a neural network-based perception pipeline).
Although these approaches show potential in efficiently generating local motion plans, they cannot enforce dynamic constraints without external help or guarantee the solutions' validity.
We propose \textit{Interpolating Radial Basis Function Networks} (IRBFN), a differentiable trajectory generation method that produces dynamically feasible trajectories and can be efficiently parallelized. Unlike existing differentiable planners, the gradient information is available throughout the process from local goal to final states on the trajectory. Our approach also leverages the GPU for highly parallelizable computations for efficiency.
One key contribution of our work is that it preserves dynamic constraints and approximates solutions from optimal control problems arbitrarily well if enough training samples are provided. Another key contribution compared to existing work is that the planner is differentiable with respect to \textit{all} possible parameters, e.g. gradients can flow from control or planning loss to the local goal selection.
The scope of this paper only includes providing a theoretical contribution that enables the use of differentiable planners and provides an important error bound. As shown in Figure \ref{fig:stack_compare}, differentiable planners blend the explainability of traditional modular planners and the scalability of end-to-end planners.

In the following sections, we'll present preliminary information on trajectory generation and Radial Basis Function Networks in Section \ref{sec:prelim}, define the interpolating RBFNs in Section \ref{sec:irbfn}, discuss the error bounds of interpolation in Section \ref{sec:errbound}, and show benchmarks in Section \ref{sec:exp}.

\section{Related Work}
\subsection{State Lattice Planners}
State lattice based motion planners based on clothoids \cite{kelly2003reactive} have been used on Autonomous Vehicles since the DARPA Urban Challenge \cite{urmson_autonomous_2008} and continue to see success in highly unstructured and competitive environments as well \cite{stahl2019}. It has also been shown that they can be easily integrated as differentiable planners \cite{karkus2022diffstack, zeng2019end}. The sampling-based scheme that state lattice planners use is flexible for planning cost evaluation and quick trajectory optimization.

\subsection{Data-driven Motion Planning}
One line of work advocates the use of a fully end-to-end trainable Autonomous driving stack and learning a network-based policy directly from a large amount of data \cite{bojarski2016end, sallab2017deep, jain2021autonomy}. These approaches boast the scalability that comes with using black-box algorithms. However, fully end-to-end approaches lack the explainability and interpretability of more principled and traditional motion planning approaches. In addition, verifiability and safety guarantees become extremely hard to inject into these algorithms due to the use of black-box approaches.
In contrast, another line of work aims to find a middle ground between end-to-end pipelines and traditional motion planning \cite{zeng2019end,zeng2020dsdnet,cui2021lookout,casas2021mp3}. These approaches try to preserve interpretability in planning by using objectives provided by the prediction and detection modules. And use a joint backbone network that takes sensor observations directly to decision-making to utilize an end-to-end model. 

\subsection{Differentiable Modeling and Motion Planning}
More recently, approaches have focused more on applying differentiable algorithms to Autonomous driving. 
The idea is to make existing model-based algorithms differentiable.
Differentiable modeling and simulation \cite{heiden2019interactive,degrave2019differentiable,song2020learning,de2018end,heiden2021neuralsim,lutter2021differentiable} has been shown to help achieve a better synthesis of controllers and improve sample efficiency using Reinforcement Learning and other learning-based methods.
Differentiable planning has shown promise in general planning task \cite{tamar2016value,nardelli2018value,lee2018gated,chaplot2021differentiable} as well as vision-based planning tasks and planning under uncertainty \cite{karkus2019differentiable,karkus2017qmdp,gupta2017cognitive}.
Lastly, differentiable control \cite{okada2017path,amos2018differentiable,pereira2018mpc,east2020infinite} has also shown performance on par with traditional control strategies and provides additional gradient information that could help improve upstream modules in the software stack.
Our approach aligns with this theme closest by making model-based algorithms differentiable. IRBFNs preserve the interpretability of planning with state lattice planners and cubic spirals. It uses a differentiable function approximator that can get arbitrarily close to the ground truth to make the planning pipeline differentiable.
Compared to approaches similar to \cite{karkus2022diffstack}, the gradient information is available for all planner parameters, e.g. gradients can flow from control or planning loss to the local goal selection policy. This enables true end-to-end training of the planner without sacrificing the properties of model-based algorithms.

\subsection{Function Approximator in Dynamics}
The method used in this paper is also related to methods proposed by \cite{ferrari2005smooth}, where ANNs are used to create differentiable function approximations, and the approximation properties are studied. Similarly, \cite{pottmann_nonlinear_1997} describes methods for characterizing non-linear plant models and controllers for these models in terms of RBFNs. However, neither of these provide an approximation error bound. Additionally, the training dataset in these works does not preserve the exact fitting of training data.

%% file: method.tex
\section{Methodology}
\subsection{Preliminaries}
\label{sec:prelim}

\subsubsection{Trajectory Generation}
\label{sec:prelimtraj}
We use clothoids as the parameterization of dynamically feasible trajectory for Ackermann steering vehicles because clothoids are posture continuous and can be represented as a polynomial. We represent the curvature of the trajectory as a cubic polynomial of the arc length $s$:
\begin{equation}
\label{eq:poly}
    \kappa(s) = a + bs + cs^2 + ds^3
\end{equation}
Following \cite{howard2009adaptive}, we re-formulate the above cubic polynomial such that the parameters are the curvatures of four equidistant points along the trajectory:
\begin{equation}
\label{eq:reformulate}
    \begin{split}
        a &= \kappa_0 \\
        b &= -\frac{1}{2}\frac{-2\kappa_3+11\kappa_0-18\kappa_1+9\kappa_2}{s_f - s_0} \\
        c &= \frac{9}{2}\frac{-\kappa_3+2\kappa_0-5\kappa_1+4\kappa_2}{(s_f-s_0)^2} \\
        d &= -\frac{9}{2}\frac{-\kappa_3+\kappa_0-3\kappa_1+3\kappa_2}{(s_f-s_0)^3}
    \end{split}
\end{equation}
To incorporate dynamic constraints into the generation process, we use the following kinematic dynamic model of the vehicle:
\begin{equation}
\label{eq:dyn}
    \begin{split}
        \mathbf{x} = \left[x, y, \theta, \kappa\right]&~~~\mathbf{u} = \left[v, \dot{\delta}\right] \\
        \dot{x}(t) &= v(t)\cos\theta(t) \\
        \dot{y}(t) &= v(t)\sin\theta(t) \\
        \dot{\theta}(t) &= \kappa(t)v(t) \\
        \dot{\kappa}(t) &= \frac{\dot{\delta}(t)}{L}
    \end{split}
\end{equation}
where $\mathbf{x}$ is the state vector consisting of the pose of the vehicle and the current curvature, $\mathbf{u}$ is the input vector consisting of the vehicle's velocity and steering velocity input, and $L$ is the wheelbase of the vehicle.
At low steering angles, we can approximate the curvature of the vehicle with the steering angle: $\kappa=\frac{\tan\delta}{L}\approx\frac{\delta}{L}$.
Additionally, we introduce constraints on the initial and final states of the trajectory:
\begin{equation}
\label{eq:cons}
    \begin{split}
        \mathbf{x}(s_0) &= \left[x_0, y_0, \theta_0, \kappa_0\right] \\
        \mathbf{x}(s_f) &= \left[x_g, y_g, \theta_g, \kappa_g\right]
    \end{split}
\end{equation}
The initial state constraint is trivial since it's determined by the current state of the vehicle. The final state is determined by the goal pose and goal curvature of the vehicle.
Next, we rewrite the formula using the substitution $v=\frac{ds}{dt}$ to find the ODEs with respect to the arc length. By dividing $v$ on both sides, the first three terms in the dynamics then become the following equations.
\begin{equation}
    \label{eq:dyn_reform}
    \begin{split}
    \frac{dx}{ds} = \cos\theta(s),~~~~~~&x(s)=\int\cos\theta(s) \\
    \frac{dy}{ds} = \sin\theta(s),~~~~~~&y(s)=\int\sin\theta(s) \\
    \frac{d\theta}{ds} = \kappa(s),~~~~~~&\theta(s)=\int\kappa(s)ds
    \end{split}
\end{equation}
Then by using Equation \ref{eq:poly} as the curvature, we perform the following integrations to find the states on the trajectory.
\begin{equation}
\label{eq:int}
    \begin{split}
        \kappa(s) &= a + bs + cs^2 + ds^3 \\
        \theta(s) &= as + bs^2 + cs^3 + ds^4 \\
        x(s) &= \int_{0}^{s_f}\cos\left(as + \frac{bs^2}{2} + \frac{cs^3}{3} + \frac{ds^4}{4}\right) ds \\
        y(s) &= \int_{0}^{s_f}\sin\left(as + \frac{bs^2}{2} + \frac{cs^3}{3} + \frac{ds^4}{4}\right) ds
    \end{split}
\end{equation}
We can then set up an optimization where the objective is to minimize the Euclidean distance between the goal pose $(x_g, y_g, \theta_g)$ of the trajectory and the integrated pose at the final arc length:
\begin{equation}
\label{eq:obj}
    \text{minimize}~~~~~|x(s_f)-x_g|^2 + |y(s_f)-y_g|^2 + |\theta(s_f)-\theta_g|^2
\end{equation}
Along with Equation \ref{eq:int} as the constraints for optimization, we can find the optimization variable $q=[\kappa_0, \kappa_1, \kappa_2, \kappa_3, s_f]$.
Following \cite{kelly2003reactive}, the position quadrature gradients and Hessians can be efficiently calculated with Simpson's rule.
Thus Newton's method can be used for optimization. Alternatively, Powell's method \cite{powell1964efficient} could also be used to find the solution without the use of gradients. To enforce dynamic constraints, we clip the curvature allowed in Equation \ref{eq:int} to the actual physical limits.

\subsubsection{Look-up Tables}
Using the optimization outlined in the previous section, a grid of local goals in the car's frame can create a look-up table (LUT) that stores the optimized parameters. However, efficient online planning using the LUT requires a high resolution of the look-up grid. In addition, points between the grid points can't be interpolated accurately using the stored values.

\subsubsection{Radial Basis Function Networks:}
 Radial Basis Function Networks (RBFNs) \cite{broomhead1988radial} use a smooth function of the distance of an input to an origin in place of a sigmoidal function as a neuron in sigmoidal neural networks. We define RBFNs following the standard definition as follows.
 
\begin{definition}[Radial Basis Function Networks]
An RBFN consists of two layers, a hidden layer with multiple RBF neurons and a linear layer. Each of the RBF neurons is centered around a predefined or trainable center where the distances are calculated.

\begin{equation}
\label{eq:rbfndef}
    \Phi(\mathbf{x}) = \sum_{i=1}^Mk_i\rho(||\mathbf{x} - c_i||)
\end{equation}
\end{definition}
$M$ is the number of centers, $c_i$s are the centers for the hidden RBF layers, $k_i$ are the weights for the linear layer. $\rho$ is the smooth activation function chosen as the Radial Basis Function. In our use case, we use an inverse quadratic function as the kernel function: $\rho(\mathbf{z}) = \frac{1}{1+z^2}$. Usually, during training, the centers of the RBFN are chosen as the available data points in the training dataset for better approximations. We left the centers as trainable parameters during our training process.

\subsection{Problem Definition}
Given a local goal $\left[x_g, y_g, \theta_g, \kappa_g\right]$ for a car-like robot, we define the trajectory generation task as creating a sequence of feasible control inputs $\left\{\mathbf{a},\mathbf{\delta}\right\}$, and corresponding poses and velocity profiles $\left\{\mathbf{x, y, \theta, \kappa}\right\}$ in the workspace, that takes the robot from its current pose to the local goal. We further formulate a parametric optimization problem to describe the generation process. From Section \ref{sec:prelimtraj}, we can fully describe a single trajectory with the polynomial parameters $\left[\kappa_0, \kappa_1, \kappa_2, \kappa_3, s_f\right]$. Thus the parametric optimization problem is:

\begin{equation}
    \label{eq:optdef}
    \begin{split}
        \text{minimize}&~~~|x(s_f)-x_g|^2 + |y(s_f)-y_g|^2 + |\theta(s_f)-\theta_g|^2 \\
        \text{subject to}&~~~ x(s) = \int_{0}^{s_f}\cos\left(as + \frac{bs^2}{2} + \frac{cs^3}{3} + \frac{ds^4}{4}\right) ds \\
        &~~~ y(s) = \int_{0}^{s_f}\sin\left(as + \frac{bs^2}{2} + \frac{cs^3}{3} + \frac{ds^4}{4}\right) ds \\
        &~~~ \theta(s) = as + bs^2 + cs^3 + ds^4 \\
        &~~~a = \kappa_0 \\
        &~~~b = -\frac{1}{2}\frac{-2\kappa_3+11\kappa_0-18\kappa_1+9\kappa_2}{s_f - s_0} \\
        &~~~c = \frac{9}{2}\frac{-\kappa_3+2\kappa_0-5\kappa_1+4\kappa_2}{(s_f-s_0)^2} \\
        &~~~d = -\frac{9}{2}\frac{-\kappa_3+\kappa_0-3\kappa_1+3\kappa_2}{(s_f-s_0)^3}
    \end{split}
\end{equation}

Then, ultimately, we frame it as a function approximation problem.
Where we approximate a function $f_{\operatorname{opt}}$ such that:
\begin{equation}
    \label{eq:optfunc}
    f_{\operatorname{opt}}\left(\left[x_g, y_g, \theta_g, \kappa_g\right]\right)=\left[\kappa^*_0, \kappa^*_1, \kappa^*_2, \kappa^*_3, s^*_f\right]
\end{equation}
Where the function being approximated takes in the local goal and outputs the optimized parameters that describe the clothoid. In the next section, we'll introduce our proposed function estimator.

\subsection{Interpolating Radial Basis Function Networks}
\label{sec:irbfn}
\begin{figure}[]
    \centering
    \includegraphics[width=\columnwidth]{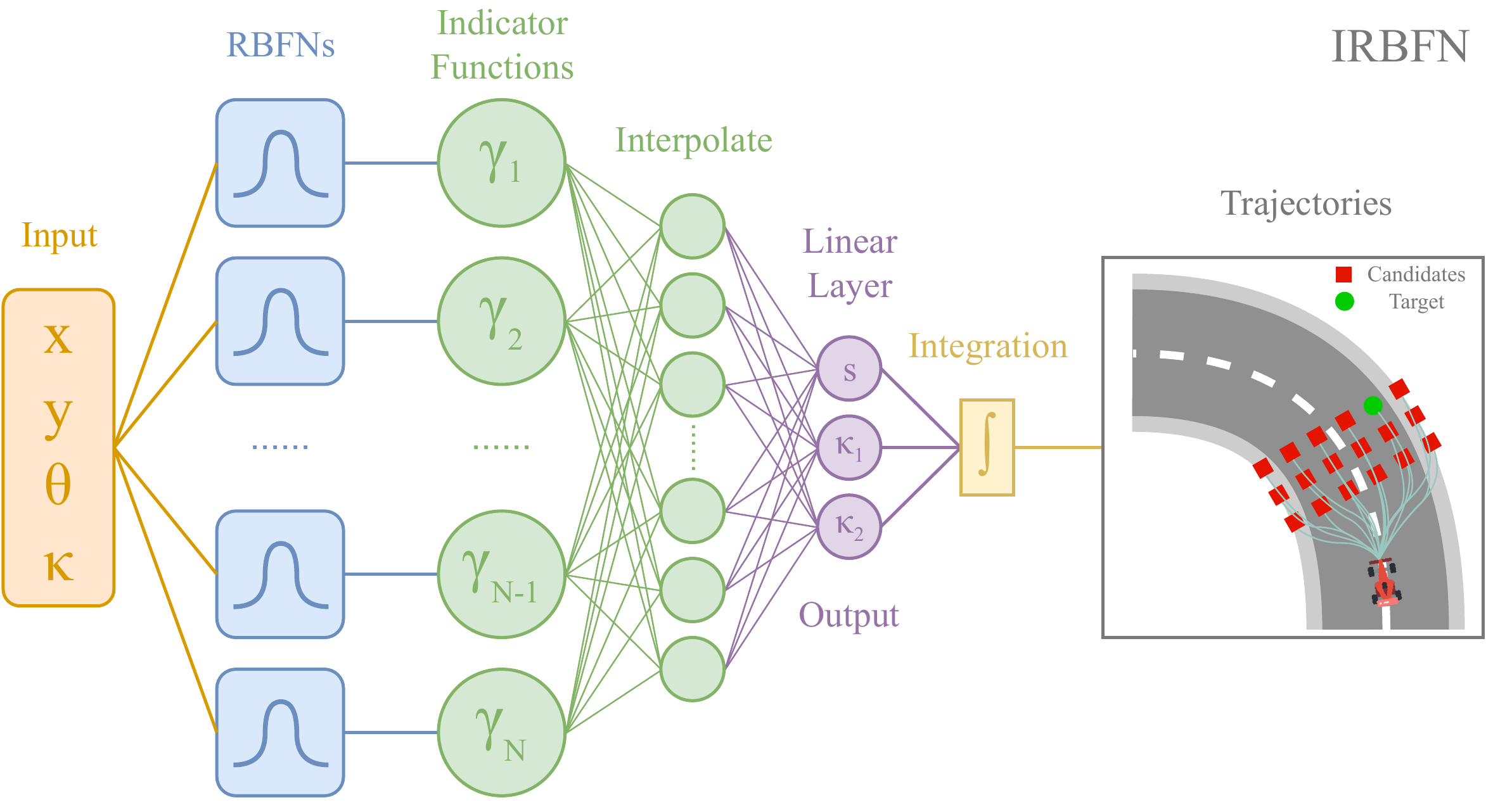}
    \caption{Network architecture of IRBFNs in the trajectory generation use case. The network takes a set of local goals as input and outputs parameters of polynomials describing the desired clothoids.}
    \label{fig:arch}
    \vspace{-10pt}
\end{figure}

We introduce a modified RBFN that performs interpolation over a uniform finite grid of the domain. Together with a smooth indicator function, interpolating approximations from multiple RBFNs produces bounded approximation error. The interpolating RBFN consists of multiple RBFNs where each RBFN performs approximation over an orthotope partitioned by a uniform grid over the domain of $f_{\operatorname{opt}}$ (shown in Figure \ref{fig:arch}).
We define a smooth and differentiable indicator function that returns a scalar between 0 and 1 on each region in each dimension of the input vector. 

\begin{definition}[Smooth Indicator Function]
    A smooth indicator function is a function $\gamma_m:\mathbb{R}^N\rightarrow[0, 1]$ defined on the orthotope $r_m=[l_1, u_1]\times\cdots\times[l_N, u_N]\in\mathcal{X}$.

\begin{equation}
\label{eq:indicator}
\begin{split}
    \gamma_{n}(\mathbf{x})=\prod_{m=1}^{M}\left(\frac{\sigma\left(\zeta\left(\mathbf{u}_{m, n}-x_{m}\right)\right)+1}{2}\right) \\
    \times\left(\frac{\sigma\left(\zeta\left(x_{m}-\mathbf{l}_{m, n}\right)\right)+1}{2}\right)
    \end{split}
\end{equation}
Where $M$ is the dimensions of the input vector, $N$ is the number of orthotopes defined, $\sigma$ is the $\mathtt{tanh}$ activation function, $\zeta$ is a scalar parameter, and $\mathbf{u}$ and $\mathbf{l}$ are the bounds of the intervals that define the orthotopes.
\end{definition}
The bounds of the intervals are defined to be multiples of $\mathbf{s}$, which is the spacing of the partitioning grid over $\mathcal{X}$. In addition, we define a controllable parameter $\delta\in[0,1]$ as a function of $N$ and $\zeta$.
An important property of the indicator function emerges by carefully choosing the values of $\delta$ and $\zeta$. At any point $\mathbf{x}\in\mathcal{X}$, the value of $\sum_{n}\gamma_n(\mathbf{x})$ can be made arbitrarily close to 1, thus creating smooth interpolation over all the orthotopes.
Intuitively, the indicator function is designed such that the output of the interpolating RBFNs at the boundaries of the orthotopes in each dimension is from exactly one-half of the outputs of each neighboring RBF. The interpolation is performed before the final linear layer in the network. Thus, we define the Interpolating Radial Basis Function Network (IRBFN) as:
\begin{definition}[Interpolating Radial Basis Function Network (IRBFN)]
    \begin{equation}
    \label{eq:irbfndef}
        \Phi_{\mathrm{interp}}(\mathbf{x}) = \sum_{i=1}^Nk_i \rho(||\mathbf{x} - c_i||)\gamma_i(\mathbf{x})
    \end{equation}
\end{definition}

\subsection{Bounded Interpolation Error}
\label{sec:errbound}
An important property of the interpolating RBFNs is bounded interpolation error. In the following section, we provide proof with the following sketch: we first show that $f_{\operatorname{opt}}$ is a computable function, hence continuous. Then we show that approximation of continuous function using interpolating Radial Basis Functions on a uniform grid provides a bounded error. First, we provide some important definitions following \cite{ko_complexity_1991}.

\begin{definition}[Oracle Turing Machine]
An Oracle Turing Machine (TM) is an ordinary Turing Machine $M$ equipped with an additional query tape and two additional states: the query state and the answer state. When the machine enters the query state, the oracle, a function $\phi$, replaces the current string $s$ in the query tape by the string $\phi(s)$, moves the tape head back to the first cell of the query tape, and puts the machine $M$ in the answer state.
\end{definition}

\begin{definition}[$k$-oracle Turing Machine]
A $k$-oracle TM $M$ is an Oracle Turing Machine which uses $k$ oracle functions $\phi_1,\ldots,\phi_k$. M can make queries to the $i$-th function $\phi_i,1\leq i\leq k$ by writing down $\langle i,n\rangle$ on its query tape and the $i$-th oracle will answer by writing $\phi_i(n)$ on the tape.
\end{definition}

\begin{definition}[Computability of functions]
A real function $f:[0,1]^k\rightarrow R$ is \textit{computable} if there exists a $k$-oracle TM $M$ such that for all $x_1,\ldots,x_k\in [0,1]$ and all $\phi_i\in CF_{x_i},1\leq i\leq k$, $M^{\phi_1},\ldots,M^{\phi_k}$ halts and outputs a dyadic rational $d$ such that $|d-f(x_1,\ldots,x_k)|\leq 2^{-n}$. Where $CF_{x_i}$ denotes the Cauchy function or the set of all functions binary converging to $x_i$.
\end{definition}

\begin{definition}[Local Lipshitz Continuity]
    A function $f$ is \textit{locally Lipschitz continuous} over a bounded domain $\mathcal{X}$ if for $\mathbf{u, v}\in\mathcal{X}$ there exists a finite $K$ such that
    \begin{equation}
        \label{eq:lipschitz}
        ||f(\mathbf{u})-f(\mathbf{v})||\leq K||\mathbf{u}-\mathbf{v}||
    \end{equation}
\end{definition}

Since $f_{\operatorname{opt}}$ generates unique solutions with gradient descent given the same initial condition in polynomial time, and the solution exists universally, $f_{\operatorname{opt}}$ is computable. Note that computable real functions were first formally defined by Grzegorczyk \cite{grzegorczyk1955computable}. The Oracle TMs definitions are equivalent to the original definition. From \cite{grzegorczyk1955computable, grzegorczyk1957definitions,ko_complexity_1991}, computable functions preserve Lipschitz continuity. Hence we can derive the following approximation bound for the interpolating RBFNs following \cite{costarelli2014sigmoidal,bejancu1999local}.

\begin{theorem}[Bounded Interpolation Error on Finite Uniform Grid]
    The maximum interpolation error is uniformly bounded by the following equation:
    \begin{equation}
        \label{eq:err}
        \begin{split}
        ||\Phi_{\mathrm{interp}}(\mathbf{x}) - & f_{\operatorname{opt}}(\mathbf{x})||_{\infty} < \frac{1}{N^\alpha} \left[L2^{\frac{\alpha}{2}+1}s^{\alpha}\right. \\
        &\left.+2^{\alpha/2}s^{\alpha}||\Phi_{\mathrm{interp}}||_{\infty}+||f_{\operatorname{opt}}||_{\infty}\right]
        \end{split}
    \end{equation}
\end{theorem}
$N$ is the number of training samples, $\alpha$ is the Hölder order, $L$ is the Hölder constant, and $s$ is the spacing between training samples, equivalent to the grid spacing we've defined. Note that when the function is locally Lipschitz continuous in our case, $\alpha=1$. It is clear that as the limit of $s$ goes to zero, implying a finer grid defined for the look-up table, the interpolation error goes to zero. By increasing the number of samples, the error can be made arbitrarily small.

\subsection{Differentiability}
Using a differentiable region indicator function (Equation \ref{eq:indicator}), the interpolating RBFN is fully differentiable. In the integration step, we use Autograd \cite{maclaurin2015autograd} to make the gradient available from the sampled local goals to the states on the trajectories. The availability of gradients through the trajectory generation pipeline in a white-box model provides benefits over black-box in Model-based Reinforcement Learning \cite{lutter2021differentiable}. Moreover, incorporating neural components significantly improves computational efficiency when synthesizing controllers for dynamic systems \cite{heiden2021neuralsim}.

%% file: exp.tex
\section{Experiments}
\label{sec:exp}

\subsection{Software Implementation and Training Dataset}
The trajectory generation pipeline is written in JAX \cite{jax2018github} and FLAX \cite{flax2020github} and can be found online at \url{https://github.com/hzheng40/irbfn}. The implementation makes use of just-in-time (JIT) compilation to speed up mathematical calculations. The model utilizes automatic vectorization (\texttt{vmap}) to create the multi-headed structure of the network. Finally, during the integration step, the implementation makes use of the Haskell-like type signature \texttt{scan} to eliminate for-loops with carryovers.
The training dataset is generated using Newton's method using the Jacobians and Hessians found in \cite{kelly2003reactive}. The resolution of the look-up table is specified in Table \ref{tab:reso}. All points are used as the training set since overfitting the available data is desired.
The interpolating RBFNs use 100 trainable centers for each RBFN and 880 regions (RBFNs). The length of the intervals that defines the orthotopes (regions) is 1.0 meter in $x$, 1.6 meters in $y$, and $0.39$ radians in $\theta$. $\zeta$ used in the indicator function are 15 for $x$, 15 for $y$, and 100 for $\theta$.
Finally, the network is trained using the Adam \cite{kingma2014adam} optimizer with a learning rate of 0.001, MSE loss, and batch size of 2000. At 400 epochs, the average training loss over the entire dataset is 0.03107.

\begin{table}[t]
    \caption{Training Data (Look-up Table) Resolution}
    \label{tab:reso}
    \centering
    \begin{tabular}{|l|c|c|}
    \hline
    Dimension  & Value & Resolution\\
    \hline
    Min $x$ (m) & 1.0 & \multirow{2}{*}{0.1}\\
    Max $x$ (m) & 10.0 & \\
    \hline
    Min $y$ (m) & -6.0 & \multirow{2}{*}{0.1}\\
    Max $y$ (m) & 6.0 &\\
    \hline
    Min $\theta$ (rad) & $-\pi/2$ & \multirow{2}{*}{0.1}\\
    Max $\theta$ (rad) & $\pi/2$ &\\
    \hline
    \multicolumn{2}{|c|}{\# Points} & 3213142 \\
    \hline
    \end{tabular}
    \vspace{-10pt}
\end{table}

\subsection{Benchmarks and Trajectory Generation Errors}
\begin{figure}[]
    \centering
    \includegraphics[width=0.9\columnwidth]{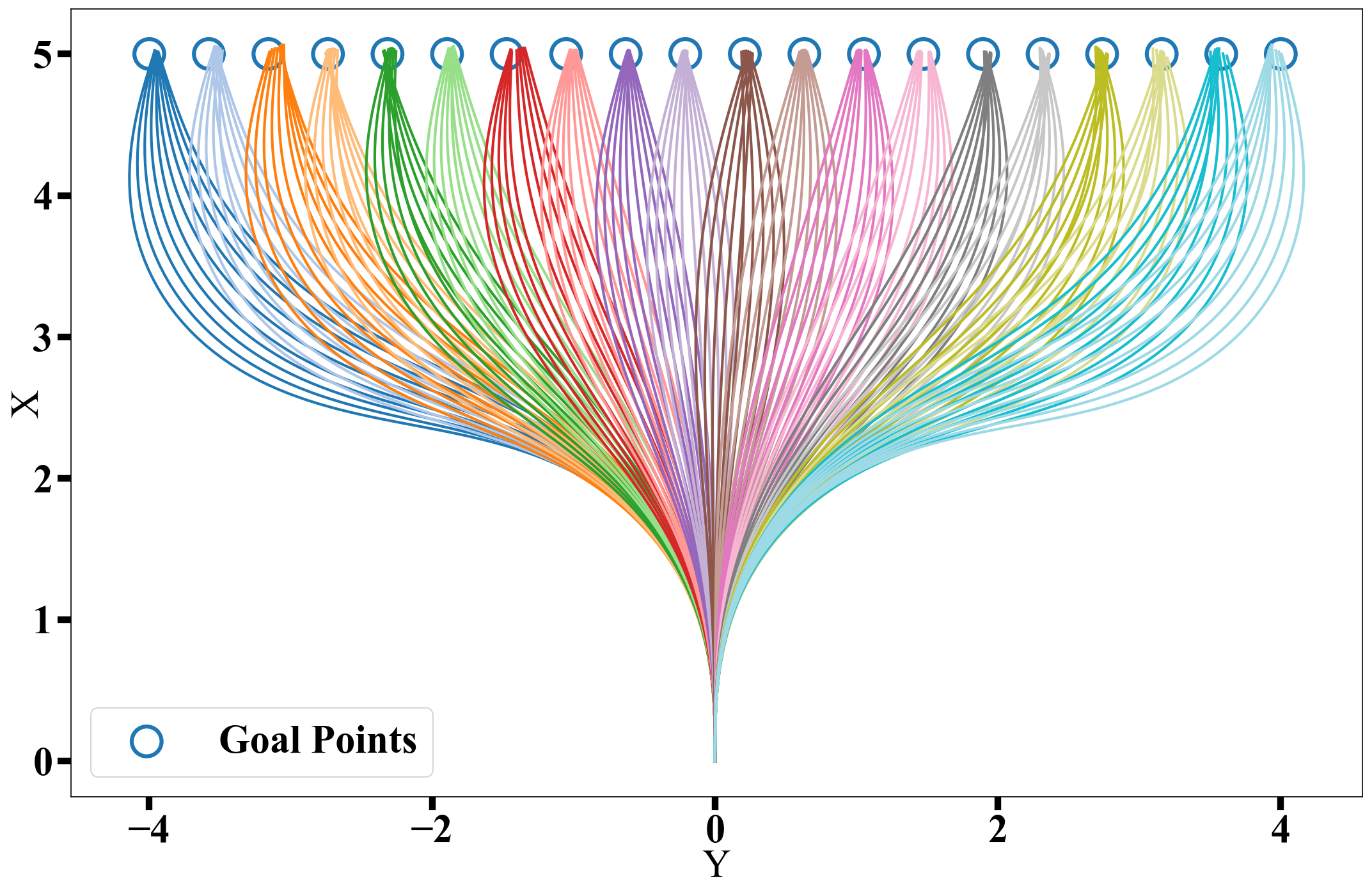}
    \vspace{-5pt}
    \caption{Example trajectory output from IRBFN at $x=5$m, and at various $y$ and $\theta$ values.}
    \label{fig:traj_out}
    \vspace{-18pt}
\end{figure}

We run all benchmarks on a system with an NVIDIA RTX 2070 Super GPU and an AMD Ryzen 9 3900X CPU. We profile the generation of 500 trajectories with different goal points.
The peak VRAM usage using our approach was 273.58 MiB. In 1000 evaluations with different random noises at each evaluation added to the goal points, our approach was able to achieve an update frequency of 230.08 Hz. Compared to optimizing for the polynomial solutions online at 3.25 Hz, our approach is a 70x+ speed up only using a small amount of VRAM. Figure \ref{fig:traj_out} shows example outputs with goals set at $x=5$ meters with various $y$ and $\theta$ values.

We also measure the average error between generated trajectories' endpoints and the given goal points. Across 500 trajectories spanning a region of 2 to 6 meters in $x$, -4 to 4 meters in $y$, and -0.3 to 0.3 radians in $\theta$. We compare the experimental error and the theoretical error bounds at the endpoints of trajectories in Table \ref{tab:err}. The theoretical error bounds are obtained by using Equation \ref{eq:err}, then propagated through the dynamics integration in Equation \ref{eq:int}. We note that while the $x$ and $y$ experimental average errors are within the theoretical bound, the $\theta$ error is above the theoretical bound. This could be due to intrinsic sensitivity in the $\theta$ dimension. Since the absolute values for $\theta$ are much smaller than those of $x$ and $y$ while having the same resolution in the look-up table used as training data. Additionally, since our training error is not zero, the prediction error adds to the interpolation error, which gets compounded through integration to generate the states on the trajectories.
Another noticeable decay in the accuracy of trajectory generation is as the desired goal moves towards the edge of the available training data.
\begin{table}[b]
    \vspace{-10pt}
    \caption{Trajectory Generation Endpoint Errors}
    \label{tab:err}
    \centering
    \begin{tabular}{|c|c|c|}
    \hline
    Dimension  & Experimental Error & Theoretical Error\\
    \hline
    $x$ (m) & 0.0264 & 0.0410\\
    \hline
    $y$ (m) & 0.0365 & 0.0410\\
    \hline
    $\theta$ (rad) & 0.0110 & 0.0008\\
    \hline
    \end{tabular}
\end{table}

%% file: discussion.tex
\section{Limitations and Conclusions}
\textbf{Limitations: }One of the limitations is the difficulty of filtering out invalid training data during the offline generation of the look-up table. In our experiments, we reject invalid trajectories by comparing the arclength and the corresponding goal's $x$ and $y$ coordinates. The percentage of valid trajectories can be improved by increasing the iteration limits in the optimization. However, this step has a tradeoff between quality and computation time.
Another limitation is that this work doesn't show the possible improvement in planning using the gradient information provided by the differentiable pipeline. Future can utilize a trainable goal selection policy and the IRBFNs in an end-to-end pipeline, i.e., model-based RL.
Lastly, calculating the gradient information is noticeably more computationally intensive than inferencing on the trained network. Future work could benchmark and improve the efficiency of gradient calculations.

\textbf{Conclusions: }In this paper, we proposed a differentiable trajectory generation pipeline for car-like robots with interpolating Radial Basis Function Networks.
Though using IRBFNs to implement a complete planner stack is out of the scope of this paper, our approach provides an important theoretical contribution that shows success in trajectory generation and a uniformly bounded interpolation error.
In terms of computation efficiency, our implementation achieves a 70x+ speed up at 230+ Hz when generating 500 trajectories simultaneously compared to existing methods while only using a small amount of VRAM.
In addition, the gradient information of every parameter is available throughout the pipeline.

%% file: ack.tex
\section*{Acknowledgments}
We thank Matthew O'Kelly for the initial ideation of the project and Joshua P. Reddy for their contribution to the initial experiments of the project.